\documentclass{article}

\usepackage[nonatbib,final]{nips_2016}

\usepackage[utf8]{inputenc} 
\usepackage[T1]{fontenc}    
\usepackage[colorlinks]{hyperref}
\usepackage{url}            
\usepackage{booktabs}       
\usepackage{amsfonts}       
\usepackage{nicefrac}       
\usepackage{microtype}      
\usepackage{graphicx, graphics}
\usepackage{caption}
\usepackage{subcaption}

\title{Alpha-Mini: Minichess Agent with Deep Reinforcement Learning}

\author{
  Michael Sun \\
  Department of Computer Science\\
  Stanford University\\
  Stanford, CA 93405 \\
  \texttt{msun415@cs.stanford.edu} \\
   \And
   Robert Tan \\
   Department of Computer Science \\
   Stanford University\\
  Stanford, CA 93405 \\
  \texttt{rtan21@cs.stanford.edu} \\
}

\begin{document}

\maketitle

\begin{abstract}
  We train an agent to compete in the game of Gardner minichess, a downsized variation of chess played on a 5x5 board. We motivated and applied a SOTA actor-critic method Proximal Policy Optimization with Generalized Advantage Estimation. Our initial task centered around training the agent against a random agent. Once we obtained reasonable performance, we then adopted a version of AlphaGo's iterative policy improvement to pit the agent against increasingly stronger versions of itself, and evaluate the resulting performance gain. The final agent achieves a near (97\%) perfect win rate against a random agent. We also explore the effects of pretraining the network using a collection of positions obtained via self-play. Code is available \href{https://github.com/shiningsunnyday/mcts-chess/}{here}.
  
\end{abstract}


\section{Problem Statement}

\subsection{Description of Problem}

Gardner Minichess is a chess variant where the board size is 5x5 and each player aims to capture the opponent's King. Standard chess rules for pieces apply. Each player (or agent) alternate turns making moves (or taking steps in the environment) with white moving first. The first player that captures the opponent's King wins. Our goal is to produce an agent that can significantly outperform the baseline (a random agent) using modern reinforcement learning techniques from self-play.

\subsection{Motivation}

Our project aims to apply decision-making techniques to playing a smaller variant of chess - Gardner Minichess. We chose this variant because:
\begin{enumerate}
    \item 5x5 is the first non-trivial board size for which no forced wins have been found.
    \item It’s the largest board size preserving the tractability of our problem given our budget\footnote{3x4 and 4x4 chess have $1.7 \times 10^{11}$ and $3.7\times 10^{15}$ positions respectively.}.
    \item We were able to reference an existing project done on meta-learning for minichess variations.
\end{enumerate}

\subsection{Environment}
In order to take advantage of RLLib and its RL algorithms, we must first define the environment for the problem. With Gardner Minichess, the state space is discrete with $5 \times 5 \times 6 \times 2 \times 2$ possible configurations as a loose upper bound (since there are 6 pieces per color with 2 colors and we must encode whose move it is). The action space is also discrete with ~1000 actions that can be taken across the board for the different number of pieces.

We have encoded the state space as a discrete multi-dimensional box and the action space as a discrete integer between 0 and ~1000, which each numerical value corresponding to a unique action (which may or may not be legal). The actions can then be coded as one-hot vectors as well.

We represent an observation as a 5x5 board and legal action mask. The 5x5 board holds as each element the value of the piece residing at that square, a positive being a white piece, its negation being the black piece. The legal action mask is used to post-normalize our policy's outputs after softmax, so the exploration settings can only select amongst legal moves. Our adoption of piece values is consistent with the \cite{mitguy} prior work. 

The true reward of the environment only occurs at a terminal state, where the win of a king corresponds to a reward proportional to the king's piece value (60000), but we have augmented the reward to capture a heuristic that allows for easier learning. Because our board directly represents pieces by their piece values, we can reward the agent at every step by the piece-value difference of its color on the board minus the piece-value difference of its color on the board in the previous step. For example, if white captures a knight or king after taking a step, its reward will be proportional to 300 and 60000 respectively.

The multi-agent environment is similar to the single-agent, except the "agent" keeps track of two policies and critics - one policy and critic for white and black each, and has to input/output two sets of {reward, done, observation} at every time-step. Our choice of implementation always represents the black agent's observation board as from white's perspective, instead relying on the second policy to associate a negative reward on the same position in training the critic. Admittedly, this may seem unconventional, having two redundant networks (why not just negative the white critic's output for black?). The reason is because we're sharing layer weights between the policy and critic, so we must have the same model output two different (possibly asymmetrical) legal action masks on the same board, and thus we needed two networks anyways, making it natural to have one network handle generating the value and policies for white and one for black separately.

\section{Introduction of Approach}
\label{gen_inst}
We aim to employ a reinforcement learning approach using the Proximal Policy Optimization (PPO) algorithm implemented in Python Ray RLLib library using Generalized Advantage Estimation (GAE). PPO is a policy gradient method and GAE is a way to estimate advantages which generalizes temporal residual difference error. The library exposes PPO amongst other SOTA algorithms, with the PPO implementation exposing many configurable modules like a custom policy and/or value model and exploration. We will train in both the single-agent environment and multi-agent environments described before, and evaluate the win rate on 10000 games against a random agent.

\subsection{Function Approximation}

We adopt a custom neural network architecture for computing both the value and policy with shared weights. Our model consists of four convolutional blocks, each with a 3x3 filter, stride=1 convolutional layer, batch normalization, and ReLU non-linearity layer each with num\_channels output channels. This consecutively downsizes the 5x5 board to a num\_channels x 1 x 1 feature map, which is flattened and passed through two consecutive fully connected + batch normalization + ReLU + dropout layers. Finally, two separate fully connected layers are applied to generate the (1,) and (|A|,) value and policy logits respectively. The policy logits is fed through a softmax and post-normalized with the action mask. 

\subsection{Exploration vs Exploitation}
A major decision we must make is controlling the extent and approach it agent takes to exploring the action space. Since we represent our policy output as a probability distribution of legal moves, we choose a stochastic sampling exploration strategy. We believe this sampling strategy both allows the agent to explore a vast state space at the start and consistently play the best moves near the end. 

\subsection{Collection of Collected Games}
A lot of work was spent bootstrapping a pretrained network using self-played games. We collected about ~12,000 completed games offline from "self-play" in order to expose the critic to a wide variety of positions and evaluations. We created training and validation sets for these over 100K positions using shallow Stockfish engine evaluation, pretrained the network to evaluate the positions, and initialized our agent's policy and value network to the pretrained network's weights in some of our runs.

\section{Relevant Literature}

Our project mainly uses PPO (Proximal Policy Optimization) in order to train a policy (agent) for both white and black using reinforcement learning \cite{schulman2017proximal}. PPO is a policy gradient method that takes estimated gradients of the policy and optimizes a clipped surrogate objective which prevents destructively large updates to the policy \cite{schulman2017proximal}, creating a pessimistic bound on the probability ratio of the policy's update. This improves over previously optimistic updates, used in trust region methods \cite{trust} and removes the need to create an additional parameter to explicitly penalize the KL divergence of the policy update via the surrogate constraint in line search \cite{mykel}, which was also shown to perform worse \cite{schulman2017proximal}. PPO (Proximal Policy Optimization) is the preferred standard baseline for many of OpenAI's projects, and is a refreshing break from traditional work for chess agents, which mostly include storing large search trees via methods like mini-max or MCTS. In particular, most modern machine learning chess AIs attempt to learn a strong evaluation function using neural networks \cite{david2017deepchess}, rather than a policy which the PPO does in parallel. GAE (Generalized Advantage Estimation) is an actor-critic method which generalizes the temporal residual difference error used for estimating the advantage, hence the gradient for the policy. It exposes a new parameter lambda to control between the length of the sequence of rollout rewards from which to estimate the advantage, effectively controlling bisa and variance. We later experiment varying lambda and gamma to gain a deeper understanding of what makes our agent successful.

\section{Training}

\subsection{Single-agent Env}
In order to train the agent, we ran Ray Tune with the PPO algorithm, which performs stochastic gradient descent on the policy at each step. We defined the training batch size to be 1000 steps (half-moves) and each iteration to terminate after 50,000 steps. For reference, each episode (game that runs from start till termination) takes around 20 steps at initialization and <10 if the agent has trained successfully. The train batch size is 1000 steps per gradient update, with each gradient update applied on SGD minibatch size of 100, with a learning rate of 1e-5 and default entropy coefficient of 0. We fix these settings for the rest of our experiments.

\subsubsection{Multi-agent Policy Improvement}

Our multi-agent training procedure has the goal of iteratively training our agent against an increasingly stronger opponent.

Initially, we initialize both a white policy $\pi_{w_0}$ and a black policy $\pi_{b_0}$ with random network weights. At every step of the procedure, we train a new policy $\pi_{w_k}$ for white against $\pi_{b_{k-1}}$ and a new policy $\pi_{b_k}$ for black against $\pi_{w_{k-1}}$. We output $\pi_{w_N}, \pi_{b_N}$ at the end, after running this for $N$ steps.

We then repeat this process several times to achieve iterative training via self-play. The same concept is used in the AlphaGo, where the agent trains against previous iterations of itself \cite{alphago}. The reason we use a distinct agent for each color is that our agents will only have been trained on one color of moves, and the action space disallows converting between one color and the other.

One slight augmentation to the iterative training that we employ is that instead of training purely against previous iterations, the agent trains against an augmented policy where with probability $\epsilon$, the policy suggests a random move. This resembles the epsilon-greedy exploration strategy that allows us to explore more unseen states even further iterations of training. We use a high $\epsilon = 0.5$ since it is very likely to encounter an unseen or underexplored state in Gardner Minichess.

\begin{figure}
  \centering
  \includegraphics[scale=0.1]{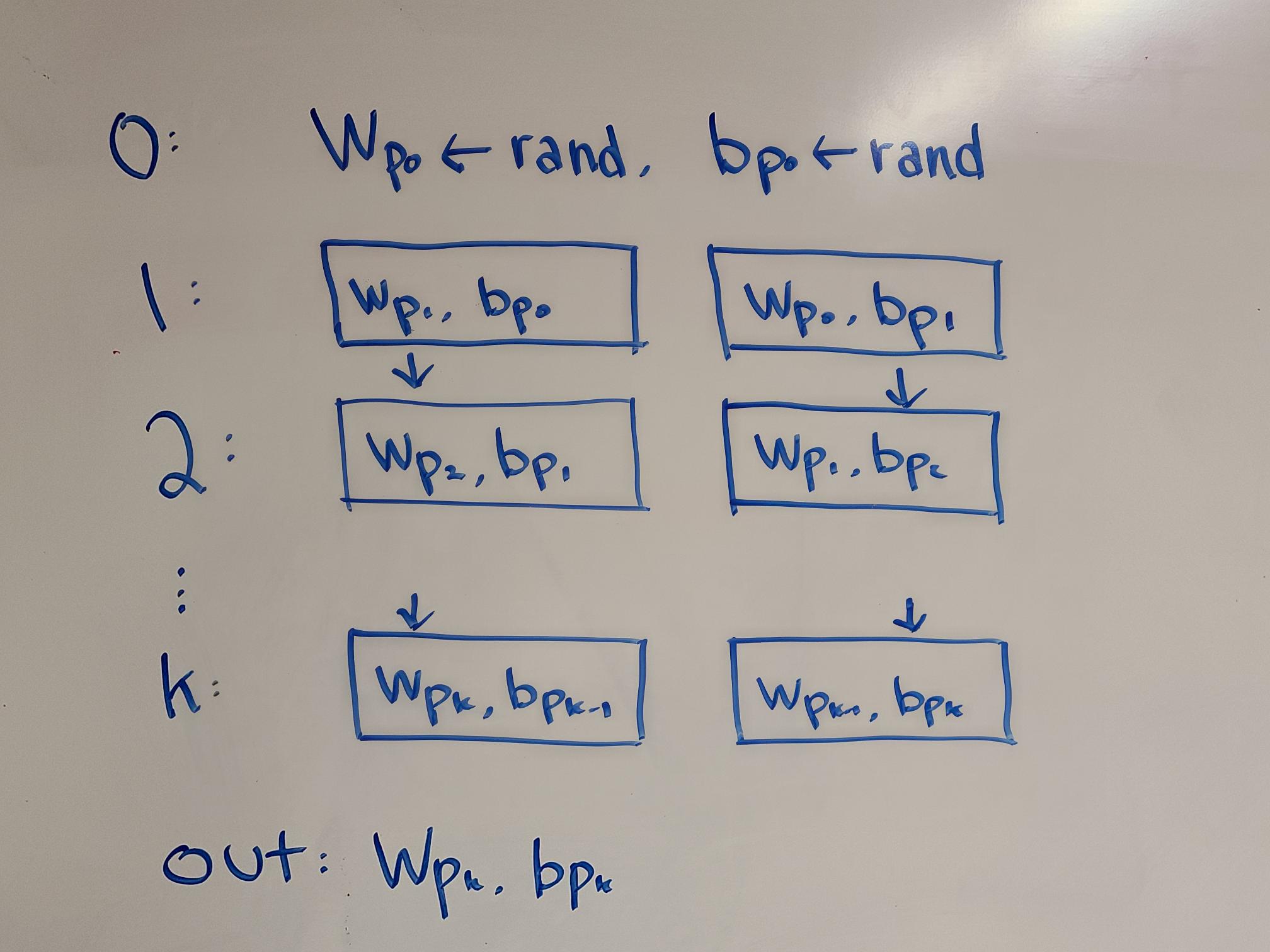}
  \caption{Multiagent Procedure}
  \label{fig:multiproc}
\end{figure}%

\section{Experiments}
\label{others}

\subsection{Evaluation Procedure}
In order to test our trained agents, we evaluate the strength of our agent against a \textbf{random} agent which just plays random (legal) moves. Our model policy is deterministic and we average over $n:= 10000$ runs due to stochasticity of the random agent, The main metric we use to monitor our experiments is mean episode reward (aka mean policy reward), which captures the average reward a policy receives across the episodes in each training iteration. With this setup, the reward (i.e. training objective) is already a near substitution to the evaluation metric, making our evaluation metric almost directly optimized over (a rare sight in ML)! This is because the policy's win rate is closely approximated just from the rewards given when a king is taken, i.e. reward ~= 60 * w + (-60)(1-w), hence win rate (w) ~= (reward + 60)/120, due to the kings having the most significant worth.

\subsubsection{Baseline (Opponent: Random Agent)}
The first such experiment is a baseline of our model against a random agent. In the random vs. random case, white won $52.7\%$ of all games (n=10000). Training against a random agent yields a $\approx 95\%$ winrate against it in our best training iteration, which is a substantial increase from the baseline model.

\subsubsection{Multiagent Policy Improvement}

For the multiagent case, we evaluate the policy by testing the white agent against our random baseline (n=10000) after every step of the policy improvement procedure and obtain the winrate.

\newpage
\section{Results}

\subsection{Single-agent Baseline}
\subsubsection{Hyperparameters}

Here, we tune the most important hyperparameters of the PPO algorithm - gamma $\gamma$ (discount factor) and lambda $\lambda$ (GAE smoothing parameter). A low $\gamma$ means that the policy will be more shortsighted and prefer actions that maximize immediate reward over long-term reward.

We have found that a low discount factor of 0.3 works best against the random agent as we have not trained for a long time and the heuristic of immediately capturing pieces performs well against a random policy, which does not always recapture pieces. Additionally, lambda of 1.0 works the best because it considers unbiased full rollouts to checkmate to estimate the advantage\footnote{That said, it does result in a higher variance, and this will become relevant in our 6.1.2 discussion on using a pretrained network.}. Since our episodes are short, the agent is able to quickly learn trajectories that capture the opponent's king.

\begin{table}[h!]
  \caption{Hyperparameter Tuning Against Random Agent}
  \label{sample-table}
  \centering
  \begin{tabular}{lll}
    \toprule
    Gamma     & Lambda     & Winrate\\
    \midrule
    0.1 & 1.0  & 0.866     \\
    0.3 & 1.0 & 0.966      \\
    0.5 & 1.0 & 0.902  \\
    0.7 & 1.0  & 0.415     \\
    0.9 & 1.0 & 0.556      \\
    & & \\
    0.3 & 0.99 & 0.948  \\
    0.3 & 0.9 & 0.853  \\
    0.3 & 0.5 & 0.691  \\
    0.3 & 0.0 & 0.687  \\

    \bottomrule
  \end{tabular}
\end{table}

\subsubsection{Pretrained Network}
We also conducted extensive experiments with a pretrained network, bootstrapped from the self-collected games.

\begin{figure}[h]
\centering
\begin{subfigure}{.3\linewidth}
  \centering
    \caption{Best Mean Policy Reward (red), with Pretrained Network (blue)}
  \includegraphics[width=\linewidth]{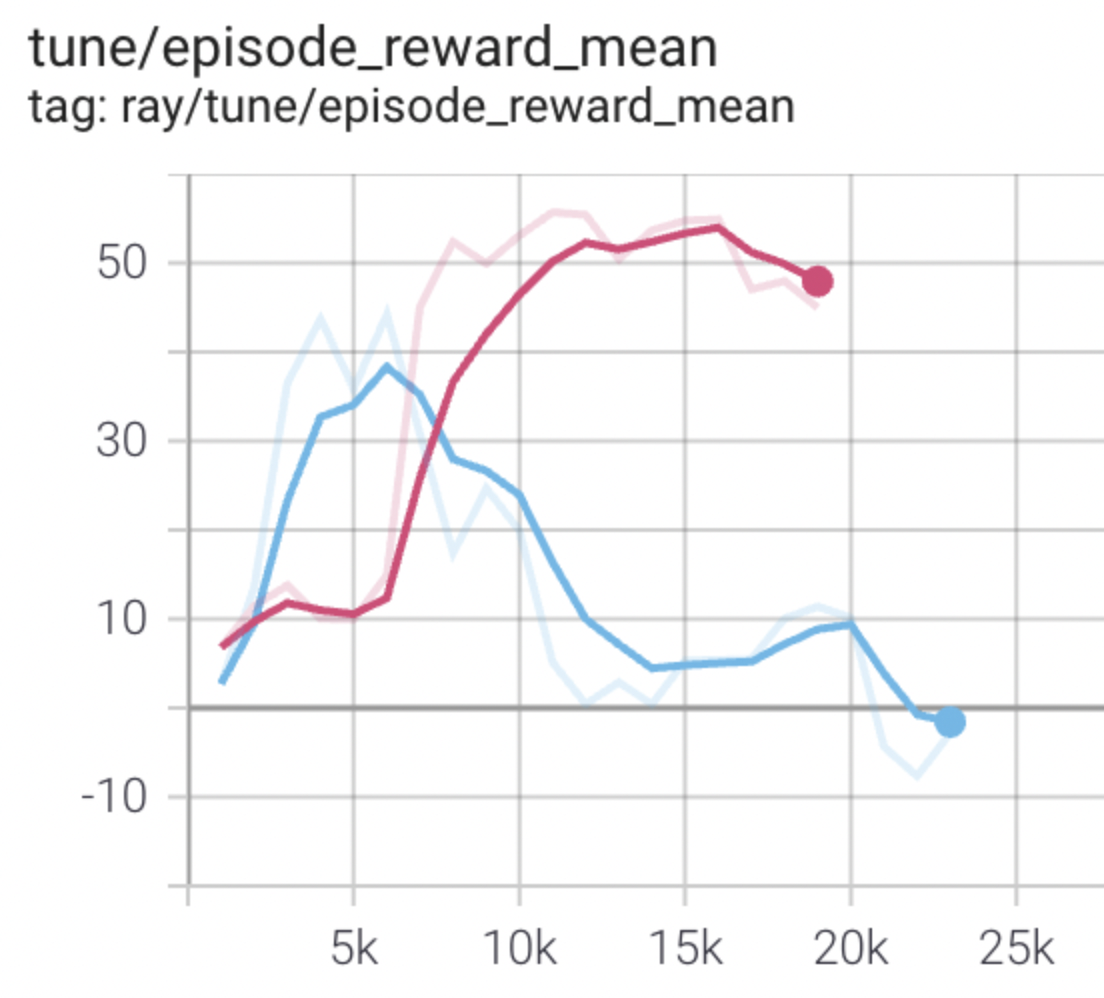}
  \label{fig:sub2}
\end{subfigure}
\begin{subfigure}{.33\linewidth}
  \centering
    \caption{Network Pretraining Each Epoch}
  \includegraphics[scale=0.2]{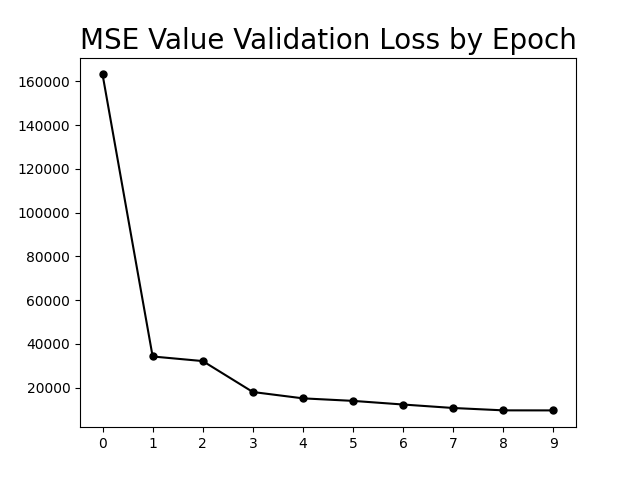}
  \label{fig:sub1}
\end{subfigure}%
\end{figure}

The pretrained network does appear to bootstrap the agent and accelerate training, with the reward curve peaking fast (in 6k timesteps as opposed to 16k) but falls fast. Despite the successful pretraining phase in which the critic was (monotonically decreasing validation loss) getting better at evaluating unseen board positions, it does not appear to help our agent. A possible explanation is that using the pretrained network initializes the agent to a local minimum, since the wide variety of positions it sees from the collected games (where random, instead of expert moves are played) likely differ from those it encounters during training. It's likely the pretrained network was not able to generalize to those positions, illustrating an issue of high variance. Given an alternative evaluation setups (i.e. move quality on boards with random piece placements), the verdict may be different\footnote{Our motivation for using a pretrained network stems from AlphaGo's success bootstrapping their network by supervising on 30 million Go positions. Without a tear drop of their computational budget, we unfortunately came up short.}. Nonetheless, we believe this is a promising direction of further research, and include this section for its practical relevance.

\subsection{Multi-agent Policy Improvement}
The following plot describes win-rates for the trained white agent at each step of the iterative training process. This shows that the iterative training process does serve to iteratively strengthen the agents that brings increasing success against the random policy up until iterations 8, after which it overfits. 
\begin{figure}[h]
  \centering
    \includegraphics[scale=0.5]{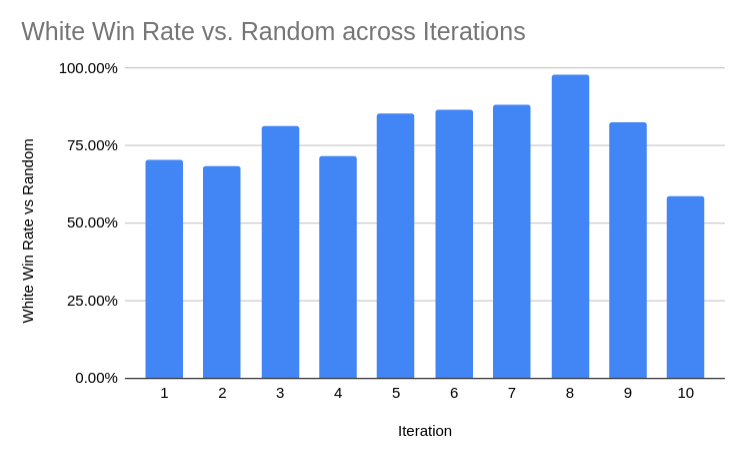}
  \caption{White agent winrates against random agent for each training iteration}
\end{figure}

\subsection{Figures}
The following figures show the mean policy reward for white ($\pi_{w_k}$) for each iteration (25k) steps of training against a fixed black agent ($\pi_{b_{k-1}}$). Recall every subsequent a new black agent ($\pi_{b_k}$) is also trained against a previous white agent ($\pi_{w_{k-1}}$) so every subsequent policy has to build on itself new strategies against a now stronger black agent, thus the reward curve has to re-climb from the bottom each iteration. From the figures, the policy reaches a peak reward between 25 and 50 before overfitting each iteration.

We have included an additional figure for the mean episode length (which is the average number of steps before reaching a terminal state) decrease. This means that through training, the agent learns methods to capture the opponents' king quicker for more immediate rewards. Each training iteration takes around 5 minutes on an RTX 2080 Super GPU.

\begin{figure}[h]
\centering
\begin{subfigure}{.3\linewidth}
  \centering
  \includegraphics[width=\linewidth]{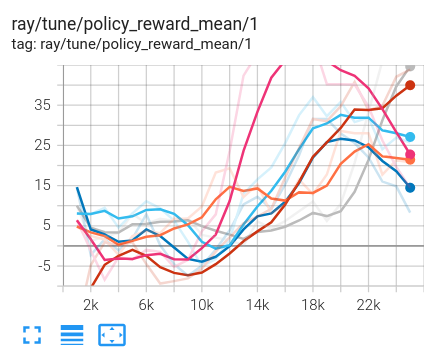}
\caption{Mean Policy Reward (Multiple Iterations)}
  \label{fig:sub2}
\end{subfigure}
\begin{subfigure}{.3\linewidth}
  \centering
  \includegraphics[width=\linewidth]{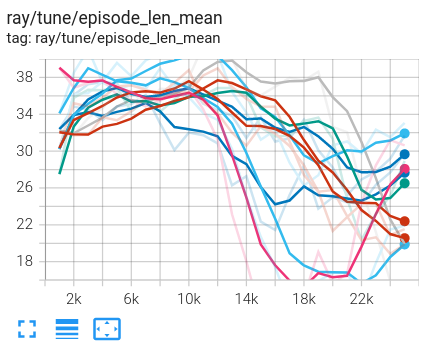}
  \caption{Mean Episode Length}
  \label{fig:sub2}
\end{subfigure}
\caption{Training White Agent Across Iterations}

\label{fig:test}
\end{figure}

\section{Conclusions}
In our experiments, we were able to train a white 5x5 minichess agent to reach over $\approx 97\%$ win-rate against a black opponent which plays random moves. We applied our knowledge of policy gradient, actor-critic and function approximation methods to employ the PPO algorithm on a downsized version of chess, avoiding using large simulation-based tree searches as used to train SOTA game-playing agents like AlphaGo/AlphaZero\cite{alphago}\cite{alphazero} or hard-coded domain knowledge like Stockfish\cite{stockfish}. We first gathered the ingredients to our solution by training a white agent in the single-agent environment, discovering the appropriate hyperparameter/training settings and exploration/reward heuristics, while exploring the effect of using a pretrained network. Following the work of AlphaGo, we then created and validated our own version of multiagent policy improvement that has the same idea of AlphaGo's self-play policy iteration while keeping consistent implementation with the action masking rules of prior work\cite{mitguy}. At every step, the white agent policy faces more and more resistance (and vice-versa). Unlike in the random, single-agent environment, the white policy now has to build off its prior knowledge and adopt new strategies to win. The resulting policy not only gets the best performance against a random agent, but is more robust to deviations in the move variations leading to checkmate.

\section{Individual Contributions}

The contributions were even across both group members. Michael worked on the single agent environment setting and the pretrained network, while Robert implemented the multi-agent setting and policy improvement procedure. Michael leveraged his prior research experience using the PPO algorithm, and Robert his past work implementing mini-max and MCTS on game-playing agents. Overall, the two partners made a great duo.

\bibliographystyle{plain}

\end{document}